\pdfoutput=1
\let\mypdfximage\pdfximage
\def\pdfximage{\immediate\mypdfximage}
\documentclass{article}

\PassOptionsToPackage{numbers, compress, square}{natbib}

\usepackage[final]{neurips_2021}

\usepackage[utf8]{inputenc} 
\usepackage[T1]{fontenc}    
\usepackage{url}            
\usepackage{booktabs}       
\usepackage{amsfonts}       
\usepackage{nicefrac}       
\usepackage{microtype}      
\usepackage{xcolor}         

\usepackage{amsmath}        
\usepackage{subcaption}
\usepackage{tikz}
\usepackage{graphicx}
\usepackage[scaled=.95]{BOONDOX-ds}
\graphicspath{{appendix/images/}{images/}}
\usepackage{ifthen}
\ifthenelse{\equal{\detokenize{main}}{\jobname}}{
  \includeonly{}
}{}
\ifthenelse{\equal{\detokenize{nohyperref}}{\jobname}}{
}{
\usepackage{hyperref}       
}

\title{Local Disentanglement in Variational Auto-Encoders Using Jacobian $L_1$ Regularization}

\author{%
  Travers Rhodes\\
  Department of Computer Science\\
  Cornell Tech, Cornell University\\
  New York, NY 10044 \\
  \texttt{tsr42@cornell.edu} \\
  \And
  Daniel D. Lee\\ 
  Department of Electrical and Computer Engineering\\
  Cornell Tech, Cornell University\\
  New York, NY 10044 \\
  \texttt{ddl46@cornell.edu} \\
}

\begin{document}

\maketitle

\begin{abstract}
  There have been many recent advances in representation learning; however, unsupervised representation learning can still struggle with model identification issues
  related to rotations of the latent space.
  Variational Auto-Encoders (VAEs) and their extensions such as $\beta$-VAEs have been shown to 
  improve local alignment of latent variables with PCA directions,
  which can help to improve model disentanglement under some conditions.
  Borrowing inspiration from Independent Component Analysis (ICA) and sparse coding,
  we propose applying an $L_1$ loss to the VAE's generative Jacobian during training
  to encourage local latent variable alignment with independent factors of variation in 
  images of multiple objects or images with multiple parts.
  We demonstrate our results on a variety of datasets, 
  giving qualitative and quantitative results using information theoretic and modularity measures that show our added $L_1$ cost encourages
  local axis alignment of the latent representation with individual factors of
  variation.
\end{abstract}

\section{Introduction}
Unsupervised representation learning takes a collection of image data from the world and figures out how to organize and find patterns in the data without additional information about how the images were generated.
The ideal representation learning algorithm would compress high-dimensional image data into a lower-dimensional latent representation that contains relevant information about the ground-truth factors of variation that generated the image.

Inferring a good latent representation from a dataset is a difficult problem and is generally underspecified in algorithms.
This underspecification is called the ``model identification'' problem, and one example is the fact that representation learning algorithms often struggle to precise the correct orientation of a latent space. 
That is, optimization criteria used to learn a representation function might be equally well satisfied by an equivalent representation that is 
just a rotation of the latent space by
an arbitrary amount.

As commonly implemented (using axis-aligned Gaussian posterior distributions),
Variational Auto-Encoders (VAEs)~\cite{kingma2014autoencoding} and their extension, the $\beta$-VAE~\cite{Higgins2017},
solve the rotational part of the model identification issue by tending to ensure that the generation function's Jacobian matrix has orthogonal columns
(i.e., that the generative Jacobian matrix's right singular values are aligned with the axes of the latent space)~\cite{Rolinek2019,pmlr-v119-kumar20d}.
Intuitively, this is because the VAE's stochastic reconstruction cost prefers to budget higher precision (lower embedding noise) in the directions along which the generative Jacobian changes most rapidly.
\citet{pmlr-v119-kumar20d} draw the parallel between this preferred orientation and linear Principal Component Analysis (PCA).
However, we note that learning algorithms that resolve rotational identification issues through methods related to PCA
will still suffer from an identifiability issue related to rotations that mix the directions for which the generative Jacobian matrix has equal singular values.
Along these directions, the posterior Gaussian would have approximately equal variance and be rotationally symmetric.

We propose adding an $L_1$ cost to the generation function's Jacobian matrix as a way to resolve that rotational identifiability issue.
Since $L_1$ cost is not rotation invariant, $L_1$ regularization creates a preferred latent-space orientation among directions whose singular values are equal.
This use of the $L_1$ norm to choose an orientation 
is inspired by similar use in linear models.
For example, when 
using a Laplacian prior in
Independent Component Analysis (ICA)~\cite{hyvarinen2000independent} and applying it to already-whitened data,
ICA rotates the data to minimize the $L_1$ norm.
As shown in~\citet{OlshausenMIT1996}, that ICA formulation is equivalent to sparse coding using an $L_1$ cost~\cite{Olshausen1996}, 
with the same rotational effect. 
In Sparse PCA, the $L_1$ norm encourages sparsity in the loadings/mixing matrix (rather than the principal components/sources as in ICA)~\cite{Jolliffe2003, Zou2006}, 
similarly encouraging preferred orientations.
These techniques for preferring certain orientations using the $L_1$ norm are all linear model techniques, and 
we apply them to non-linear VAEs by regularizing the $L_1$ norm of the generator Jacobian matrix, 
thereby encouraging a preferred orientation for our non-linear models.

\begin{figure}[ht]
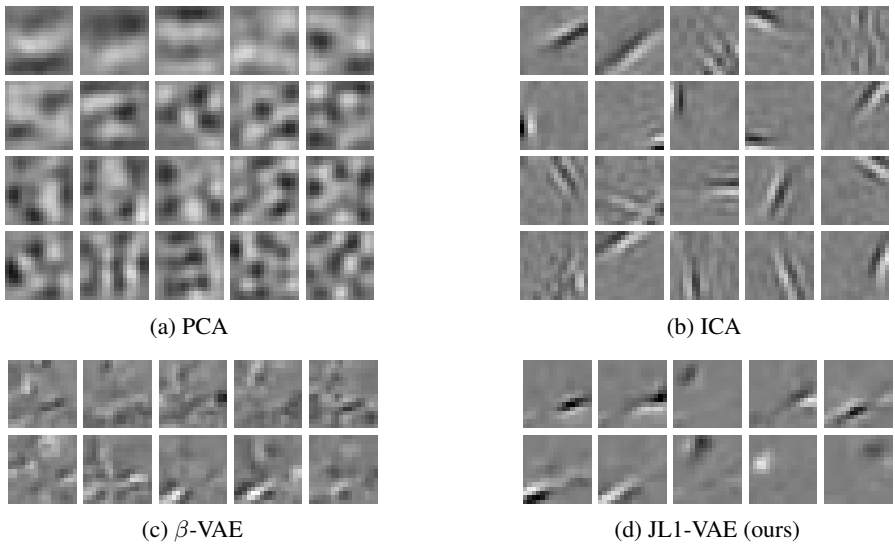
%
\centering%
\setcounter{imcnt}{0}%
\begin{subfigure}[]{0.49\textwidth}%
\centering%
\begin{tikzpicture}
    \foreach \y in {0,...,3} {
      \foreach \x in {0,...,4} {
        \node[inner sep=0] at (\x, -\y)
          {\includegraphics[width=0.9cm]{copiedAppendix/analyticModels/analyticNaturalImage/PCA_lat100_image\theimcnt}};
        \stepcounter{imcnt}
      }
    }
    \end{tikzpicture}%
\caption{PCA}%
\end{subfigure}%
\setcounter{imcnt}{0}%
\begin{subfigure}[]{0.49\textwidth}%
\centering%
\begin{tikzpicture}
    \foreach \y in {0,...,3} {
      \foreach \x in {0,...,4} {
        \node[inner sep=0] at (\x, -\y)
          {\includegraphics[width=0.9cm]{copiedAppendix/analyticModels/analyticNaturalImage/ICA_lat100_image\theimcnt}};
        \stepcounter{imcnt}
      }
    }
    \end{tikzpicture}%
\caption{ICA}%
\end{subfigure}%
\medskip

\noindent 
\def \yimcropstart{100}%
\def \ximcropstart{45}%
\def \gammaval{00}%
\setcounter{imcnt}{0}
\begin{subfigure}[]{0.49\textwidth}%
\centering%
\begin{tikzpicture}
    \foreach \y in {0,...,1} {
      \foreach \x in {0,...,4} {
        \node[inner sep=0] at (\x, -\y)
         {\includegraphics[width=0.9cm]{copiedAppendix/latentJacobianImages/naturalImageNorm/beta0_010_ica0_0\gammaval_lat10_im0_latind\theimcnt_x\ximcropstart_y\yimcropstart}};
        \stepcounter{imcnt}
      }
    }
\end{tikzpicture}%
\caption{$\beta$-VAE}%
\end{subfigure}%
\def \gammaval{10}%
\setcounter{imcnt}{0}%
\begin{subfigure}[]{0.49\textwidth}%
  \centering%
  \begin{tikzpicture}%
    \foreach \y in {0,...,1} {
      \foreach \x in {0,...,4} {
        \node[inner sep=0] at (\x, -\y)
       {\includegraphics[width=0.9cm]
         {copiedAppendix/latentJacobianImages/naturalImageNorm/beta0_010_ica0_0\gammaval_lat10_im0_latind\theimcnt_x\ximcropstart_y\yimcropstart}};
      \stepcounter{imcnt}
      }
    }
  \end{tikzpicture}%
  \caption{JL1-VAE (ours)}
\end{subfigure}%
\medskip

\noindent 
\caption{Example columns from generative Jacobian matrices for
different modeling techniques on natural image data collected in~\cite{Olshausen1996}.
ICA used 100 latent dimensions, of which 20 samples are shown. $\beta$-VAE used $\beta=0.01$; JL1-VAE used $\beta=0.01$, $\gamma=0.01$, each on ten latent dimensions.
}
\label{fig:naturalImageJacobianColumns}
\end{figure}

The motivation above suggests that an $L_1$ norm on the generator Jacobian can address the rotational identifiability issue,
and we think the $L_1$ regularization will encourage useful orientations of the latent space for image data.
This belief comes from
results on the sparse linear coding of images.
In particular, \citet{Olshausen1996} showed how sparse linear coding on natural images
generates local receptive fields
similar to those discovered in the mammalian visual processing system.
These types of localized receptive fields are shown in Figure~\ref{fig:naturalImageJacobianColumns}.
Each image in that figure corresponds to a direction in the latent space, and shows how perturbing the latent value in that direction changes the generated image---white pixels means the image gets brighter there, black pixels means the image gets darker.
For the ICA basis, we see that perturbations in different latent directions are associated with localized changes to the image, 
whereas for the PCA basis, the latent directions tend to affect the entire image.
Adding an $L_1$ penalty to our model should encourage sparsity within the generative Jacobian columns of our model, 
meaning that perturbations of latent values along individual latent directions should modify as few pixels as possible,
leading to latent directions that affect the output image in localized regions.
$L_1$ generative Jacobian regularization should therefore disentangle representations of different objects in an image.
We call the proposed model trained with this additional regularization a Jacobian $L_1$ Regularized Variational Auto-Encoder (JL1-VAE).

\newcounter{imcnt}

The use of the generative Jacobian in the JL1-VAE implies a local linear approximation of the generative function, 
so in order for the JL1-VAE to be useful,
the training image data should lie on a manifold~\cite{manifoldPerceptionLee}, 
and should be sufficiently well sampled.
Additionally, JL1-VAE contains inductive bias, like every other unsupervised disentanglement algorithm~\cite{locatello2019challenging}.
As mentioned above, JL1-VAE's regularization of the generative Jacobian encourages small changes in latent values to result in sparse (impacting a small number of pixels) changes  to the resulting image,
similar to local receptive fields.
This inductive bias is well suited for disentangling motions of different objects in an image, but would presumably not be useful for whole-image changes,
such as rotation of the entire image or brightness changes across the whole image.

We apply our novel JL1-VAE framework to a variety of datasets, 
  giving qualitative and quantitative results showing that our added $L_1$ cost can encourage 
  local alignment of the axes of the latent representation with individual factors of
  variation.

\section{Background}
\label{sec:vaeIntro}
We present a brief overview of VAEs and introduce the notation we will use throughout this paper. 
VAEs~\cite{kingma2014autoencoding} train a model to generate a data distribution that approximately matches the distribution of unlabeled training data.
For the VAEs we consider in this paper,
generating a datapoint $\tilde x \in \mathbb R^n$ from a trained VAE consists of sampling a latent variable $z \in \mathbb R^l$ from a standard multivariate Gaussian distribution $N(\mathbb{0},\mathbb{1})$ and
applying a generator function $g: \mathbb R^l \to \mathbb R^n$ to $z$ to map the latent variable to a generated image $g(z)$.
Around this generated image we assume a Bernoulli probability of similar images $\tilde x \sim p(\tilde x; g(z))$ with the generated image $g(z)$ as its mean.

To train the VAE we also define a multi-variate Gaussian embedding distribution $q(z | x)$ for each training image $x$ with mean $h(x)$ (using a learned embedding function $h:\mathbb R^n \to \mathbb R^l$) and diagonal covariance $\Sigma_{z|x}(x)$ (using a learned covariance function $\Sigma_{z|x} : \mathbb R^n \to \mathbb R^{l}$ that computes the diagonal elements).
This embedding distribution $q(z | x)$ is motivated by a desire to approximate the posterior distribution $p(z|x)$.

The objective during training of the VAE is to maximize the Evidence Lower BOund (ELBO), which is defined as $\sum_x L(x)$, where, for each data point $x$,
\begin{equation}
  L(x) = E_{z\sim q(z|x)}\left[p(x; g(z))\right]- \text{KL}\left(q(z|x)\| N(\mathbb{0},\mathbb{1})\right)
\end{equation}

The ELBO is a lower bound on $\log(p(x)) = \log(\int_z p(x|z)p(z)dz)$, which is the likelihood of the data point given our model.
Thus, maximizing the ELBO is a proxy for maximum likelihood estimation.
The $\beta$-VAE~\cite{Higgins2017} is an extension to the VAE that multiplies the second term in the ELBO by an adjustable hyperparameter $\beta$.
The first term in the ELBO is a stochastic reconstruction loss, so we sometimes refer to $\Sigma_{z|x}$ as the ``embedding noise,'' since it adds noise to the embedding when estimating the stochastic reconstruction loss. 

\citet{pmlr-v119-kumar20d} show that the stochastic reconstruction term of the ELBO, $E_{z\sim q(z|x)}\left[\log(p(x; g(z)))\right]$, can be approximated using a second-order Taylor expansion as:
\begin{equation}
  \label{eq:deterministicReconstructionApprox}
  E_{z\sim q(z|x)}\bigg[\log(p(x;g(z))\bigg] \approx \log p(x|h(x)) + \frac 1 2 \text{tr}\left(J_g(h(x))^\top H_{p_x}(g(h(x)))J_g(h(x))\Sigma_{z|x}(x)\right)
\end{equation}
$J_g$ is the Jacobian of the generator function, and $H_{p_x}(g(h(x)))$ is the Hessian with respect to $g(z)$ of the log of the generative probability $\log(p(x; g(z)))$ evaluated at $g(h(x))$. 
For standard VAE implementations using diagonal Gaussian posteriors and
pixel-factorized generative probabilities, $\Sigma_{z|x}(x)$ and $H_{p_x}(g(h(x)))$ are both diagonal.

Equation~\ref{eq:deterministicReconstructionApprox}
shows that the stochastic reconstruction loss of the ELBO can be approximated by a deterministic reconstruction loss
with a
(weighted) $L_2$ regularization cost on the Jacobian $J_g(h(x))$.
\citet{pmlr-v119-kumar20d} use this weighted $L_2$ regularization in the approximation above to show how the ELBO encourages local alignment of the right singular vectors of $J_g$ to $\Sigma_{z|x}(x)$.
That is, larger values of $\Sigma_{z|x}(x)$ (larger embedding noise) in some directions leads to larger $L_2$ regularization on the generator Jacobian in those directions. 
Equivalently, in directions with large changes in the generator Jacobian, the regularization encourages smaller embedding noise (more precision) in the posterior $\Sigma_{z|x}(x)$.
For this work, we use the presence of an implicit $L_2$ loss on the generator Jacobian
as further motivation for our choice to add an explicit $L_1$ regularization to the generator Jacobian.

\section{Related Work}
There are several areas of related work we wish to call to the reader's attention.
We note previous work in the analysis of the disentanglement properties of $\beta$-VAEs.
We delve into previous uses of the term ``sparse VAE,'' as there are (at least) two other common and well studied meanings of ``sparse VAE,'' which we disambiguate from the type of sparsity we study in this work.
We note previous VAE work inspired by ICA.
We discuss architectural choices that have been shown to improve disentanglement.
Finally, we discuss modifications to the VAE objective that have previously been studied to improve
disentanglement.

\paragraph{Disentanglement of $\beta$-VAEs}
\citet{Mathieu2019} and \citet{Rolinek2019} 
show that restricting the posterior covariance to diagonal (called the mean-field assumption) 
breaks the rotational symmetry of $\beta$-VAEs.
\citet{Rolinek2019}
further show that this encourages the columns of the generator Jacobian to be
orthogonal, relating local $\beta$-VAE latent directions with PCA decomposition.
In our work, we take this a step further, explicitly regularizing the sparsity
of the generator Jacobian, to break rotational symmetry between directions with equal singular values in of the generator Jacobian.

\paragraph{Sparse VAEs: Sparsity in VAE codes}
Some previous work involving $L_1$ regularization and VAEs uses the term ``sparse VAEs'' to refer to sparsity in the \emph{latent values} taken on by the latent codes themselves. 
That is, these works attempt a minimization of something like $\|z\|_1$. This meaning is studied in~\cite{Makhzani2014,SparseAutoEncoderNg,jiang2021improved}. When we use sparsity in the present work, however, we are not concerned with the values of the latents $z$, but rather with how small axis-aligned changes in the latent values affect the output. That is, we are concerned with sparsity of $J_g(z)$, not $z$. Our associated cost is $\|J_g(z)\|_1$. Our use of ``sparsity'' has to do with local disentanglement, rather than sparsity in latent values, which is a type of global disentanglement.

\paragraph{Sparse VAEs: Sparsity in Network Weights}
Likewise, other work involving $L_1$ regularization and VAEs (and neural networks more broadly) uses ``sparsity'' to refer to the desire to make many network weights 0, with the motivation of reducing the size of the stored neural network architecture.
This meaning is seen in~\cite{Louizos2018, Denil2013}.
In this work, by contrast, we are interested in sparsity in the generator Jacobian, not in the network weights. We note the distinction between sparsity of individual network weights and sparsity in the Jacobian of the overall function. Multiplying sparse matrices does not necessarily result in sparse matrices, and non-linear activations can allow a sparse Jacobian even if the network weights themselves are not particularly sparse. Thus, the $\|J_g(z)\|_1$ cost we study in this work is not what is referred to in studies of sparsity of neural networks, which consider regularizations like the $L_1$ cost over network weights.

\paragraph{ICA within the VAE Literature}
Independent Component Analysis (ICA)~\cite{hyvarinen2000independent} 
is often mentioned in the VAE literature in
relation to the role ICA has played in the theory of identifiability and
disentanglement of
representations~\cite{pmlr-v119-kumar20d, Mathieu2019, Rolinek2019, Kim2018, pmlr-v119-locatello20a}.
\citet{pmlr-v108-stuehmer20a} propose using a structured,
rotationally asymmetric prior
to encourage disentanglement in the embedding.
This, and other approaches that attempt to globally match the embedding
distribution to a desired shape, are very different from the
local, Jacobian-based approach we take in this paper.
\citet{pmlr-v108-khemakhem20a} 
relate nonlinear ICA with VAEs
in the case where the data has an additionally observed variable,
and they give a proof that in that case their model 
is identifiable and correctly disentangles the ground-truth factors of
variation.
We focus on fully unsupervised training data and assume that we
are not given access to any data labels.
We are not aware of any prior work 
applying a sparsity cost to the generator Jacobian, which is the inspiration we take from ICA.

\paragraph{Architectures shown to improve disentanglement}
Previous work has shown impressive results from modifying the network architecture
in order to explicitly represent multiple objects by, for example, learning
object masks~\cite{pmlr-v97-greff19a}, or by
modifying how the latent variable is read in to the
generative model architecture~\cite{watters2019spatial}.
\citet{Ainsworth2018} and \citet{khan2021} both construct model architectures that explicitly encourage sparsity in the generative network weights.
In this work, we focus on how we can regularize the
objective function to improve disentanglement, rather than studying how different network architectures can improve
disentanglement.

\paragraph{Modification of VAE prior}
Previous work has also investigated modifications to the unit Gaussian prior commonly used in VAEs. \citet{Tomczak2017} use a learnable Gaussian mixture prior.
\citet{pmlr-v108-stuehmer20a} use a generalized Gaussian distribution (that is not rotationally invariant) as the prior.
\citet{Bauer2018} use rejection sampling to form a more complicated, non-rotationally-invariant prior.
\citet{Davidson2018} and \citet{Reyb} even modify the prior to lie on non-Euclidean surfaces.
\citet{Kim2018} do not explicitly enforce a prior distribution, but rather use a regularization term to encourage the prior distribution $q(z)$ to be a factorized distribution.
While these approaches pressure the entire embedded distribution to have certain properties, we are instead focused on how to modify the learning objective to give \emph{local} bias toward disentanglement, rather than using more global methods based on the overall distribution of the embedded dataset.

\paragraph{Regularization of VAEs}
Several previous works explicitly or implicitly use $L_2$ normalization of the generator Jacobian~\cite{pmlr-v119-kumar20d, Rifai2011,Varga2018,hoffman2019robust}.
\citet{Chen2020} regularize by $\|J_g^\top(z)J_g(z) - c\mathbb{1}\|_2$ for some constant $c$.
We are not aware of previous investigations of $\|J_g(z)\|_1$ regularization for VAEs.

\section{Model Loss Calculation and Architecture}
\subsection{Loss Calculation}
We define a Jacobian $L_1$ Regularized Variational Auto-Encoder (JL1-VAE) as a
VAE that is trained using the $\beta$-VAE loss augmented with an $L_1$
regularization of the Jacobian matrix of the map from
latent values to mean generated images.
The regularization term is modulated by a hyperparameter $\gamma$.

Specifically, the maximization objective for the JL1-VAE is the sum over all datapoints $x$ of
\begin{equation}
  L_{\text{JL1}}(x) = E_{z\sim q(z|x)} \bigg[\log p(x|z) - \gamma \big|\big|J_g(z)\big|\big|_1\bigg] - \beta \text{KL}\left(q(z|x)\|N(\mathbb{0},\mathbb{1})\right) 
\end{equation}

Since we use a Gaussian posterior $q(z|x)=N\left(h(x), \Sigma_{z|x}(x)\right)$, we can use an explicit calculation
for the KL-divergence. We estimate the expectation of $\log p(x|z)$ and of $\gamma\|J_g(z)\|_1$ using a 
single sample $z$ from the distribution over which we are taking the expectation.
We estimate the full Jacobian matrix $J_g(z)$ using the finite difference method
along each latent dimension.
This leads to a
runtime that scales roughly linearly with the number of latent
variables in the VAE architecture.

\subsection{Architecture}
We use a convolutional architecture for our VAEs. 
In particular, 
our embedding architecture consists of convolutional layers followed
by a fully connected layer with ReLU activations.
This base model is shared between the mean and log variance embedding networks.
Each embedding network then appends its own linear fully connected head to the shared model.
We use a diagonal structure for the log variance
estimates to reduce the number of parameters we need to estimate.
We use a latent dimension of ten for all experiments, though we have seen similar results for other latent dimension sizes.
The reconstruction architecture consists of fully connected layers followed by
convolutional layers, using ReLU activations, with a final sigmoidal
activation function.
A full set of hyperparameter choices for each experiment can be found in the Appendix. When we compare JL1-VAE with other methods, we ensure consistent architecture choices. 

\section{Experiments}
\subsection{Datasets}
To evaluate the ability of JL1-VAE to locally disentangle factors of
variation, we apply it to a variety of datasets.

The first are natural images in grayscale taken by~\citet{Olshausen1996} and cropped to $16\times16$-pixel regions. 
We do not have labeled ``ground-truth factors of variation'' for this data, but we are able to provide qualitative results by inspecting the columns of the generator Jacobian matrix.
This data was made publicly available without a specific license, so we analyze it
under fair use.

The second is a dataset of simulated $64\times 64$-pixel grayscale images of three black dots on a white
background, inspired by~\cite{Zhao2018}.
The ground-truth factors of variation for this dataset are the x/y coordinates of the dot centers.
We re-implement \citet{Zhao2018}'s code to generate the dot images and modify the code so that dots can overlap.
We note for this dataset that if a model were to disentangle 
individual dot motions into different latent
directions, then, by symmetry, we would expect 
identical singular values of $J_g$ for those directions.
Thus, we expect that for this dataset
$\beta$-VAEs will be unable to isolate individual dot motions, 
since it has trouble disentangling directions in which the generator Jacobian has equal singular values.

Finally, we also apply our approach to tiled images of a real robotic arm taken from the
MPI3D-real dataset~\cite{Gondal2019}, licensed under Creative Commons Attribution 4.0
International License.
For each data point, we downsample four random images of the robot arm holding a
large blue square in different locations and tile the random images in a
2$\times$2 pattern to generate a new, more complicated 64$\times$64-pixel
image containing four different images of a real robotic arm.
We call this tiled image dataset MPI3D-Multi.

\subsection{Training}
For the three-dots and MPI3D-Multi datasets, we train using a Bernoulli loss on batches of 64 images over a total of 300,000
batches.
We use the Adam optimizer with a learning rate of 0.0001 (matching~\cite{locatello2019challenging}).
We use linear
annealing from 0 to the final hyperparameter value over the first 100,000 batches
for both the beta hyperparameter and JL1-VAE's $\gamma$ parameter in our implementations for JL1-VAE and $\beta$-VAE (unlike~\cite{locatello2019challenging}).
We note annealing to be beneficial to avoid model collapse when adding our $L_1$ regularization term.
We train these models on a Nvidia Quadro V100 hosted locally and one hosted on Google Cloud. 
Training each JL1-VAE model on ten latent variables
takes approximately 2.5 hours, while training each $\beta$-VAE model takes
approximately 45 minutes.
In total, training ten JL1-VAE models and ten $\beta$-VAE models for
quantitative evaluation takes roughly 33 hours.

For the natural image dataset, we train for 100,000 batches of 128 images. We use the Adam optimizer with a learning rate of 0.001 and train on a Nvidia Quadro V100s hosted locally. Training takes nine minutes for the $\beta$-VAE and 23 minutes for the JL1-VAE.

\subsection{Evaluation Metrics}
There are several metrics commonly used to measure ``disentanglement'' of a latent representation.
In this work we address two common metrics, the Mutual Information Gap (MIG)~\cite{chen2018isolating} and Modularity~\cite{ridgewayNeurIPS2018},
and show how we are able to provide extensions to these metrics that give
a measure of how well a representation \emph{locally} disentangles factors of variation.

The original MIG and modularity metrics measure global disentanglement\textemdash that is, they measure across the whole dataset how well each latent variable maps to a unique ground-truth
  factor of variation.
Since the JL1-VAE does not add an explicitly global disentanglement incentive to $\beta$-VAEs,
but instead is designed to locally encourage disentanglement using the Jacobian of the generative map,
we do not expect it to necessarily improve the global disentanglement of factors of variation.
For example, the JL1-VAE may assign factors of variation to latent
variables using one pairing in one local region of the latent space and a different pairing in a different region of the latent space.
This could lead to good average local disentanglement, but would not lead to good global disentanglement.

We are therefore interested in defining \emph{local} disentanglement metrics based on MIG and modularity.
We call these metrics ``local MIG'' and ``local modularity,'' and the general form of their calculation is
  to compute each metric 
  several times on different random ``local'' samples from the global dataset and then average the results.
 As with MIG and modularity, the calculation of these disentanglement metrics requires a generative model of the data from ground-truth factor values.

The key technique for each of our local metrics is to repeatedly compute the disentanglement metric on random local samples of data.
To generate a random local sample of data, we 
we randomly choose a centroid from the ground-truth factor values
and then random sample N ground-truth datapoints within an $L_\infty$ distance $\rho$ from that centroid.
The radius $\rho$ is a hyperparameter determining how close ground-truth factors of variation need to be in order to be considered ``local.''
We scale the hyperparameter $\rho$ as a fraction of the total range of available values for each latent variable. 
This set of N datapoints comprises each local data sample.
For our experiments we choose $N=10,000$.

For each local sample of data, we apply the MIG and modularity metrics to that sample to determine the disentanglement of the latent space in that local region. We use the MIG and modularity calculation implementations from the
open-source (Apache License 2.0) \texttt{disentanglement\_lib}
library~\cite{locatello2019challenging}.

We repeat this algorithm with 20 different local data samples and report the average as the local disentanglement score.

\section{Results}
\label{sec:Results}
We present
qualitative results for the three-dots, MPI3D-Multi, and natural image datasets, and quantitative results for the three-dots dataset.

\subsection{Qualitative Results}
Qualitative results are generated by inspecting the generator Jacobian at the (deterministic) latent embeddings for example images.
Each generative Jacobian matrix column is associated with a latent direction and shows how the generated image would change from a slight perturbation to the embedding
in that latent direction.
The generative Jacobian matrix columns for natural images are shown in Figure~\ref{fig:naturalImageJacobianColumns}.
There, we show the largest 20 PCA components, an arbitrary sample of 20 ICA components
(from training 100 ICA components using FastICA~\cite{scikit-learn}), and the
ten Jacobian matrix columns for the $\beta$-VAE and JL1-VAE models.
Additional visualizations are included in the Appendix.
We discern more localization (results more similar to local receptive fields) in the JL1-VAE and ICA results, compared to the $\beta$-VAE and PCA results.

For the three-dots dataset and MPI3D-Multi, 
we follow the same procedure to generate qualtitative results.
We show results for the three-dots dataset in Figure~\ref{fig:jacobianVisualizationThreeDots}.
There, we show the six Jacobian columns with the largest $L_2$ norms for the three-dots dataset 
 for both our JL1-VAE and $\beta$-VAE
(all ten Jacobian columns are visualized in the Appendix).
Qualitatively, we see that, when evaluating the Jacobian of the generator
function for our JL1-VAE,
individual dot motions are separated into different latent components.
The $\beta$-VAE does not exhibit this behavior. 

For the MPI3D-Multi dataset, containing tiled images of a real robot, we again
see that the JL1-VAE does a better job separating the four robots contained in
each image into separate latent variables. These results are presented in
Figure~\ref{fig:jacobianVisualizationMPI3DMulti}.

\def\boxscale{1.1}%
\def\imscale{1.0}%
\begin{figure}[t]
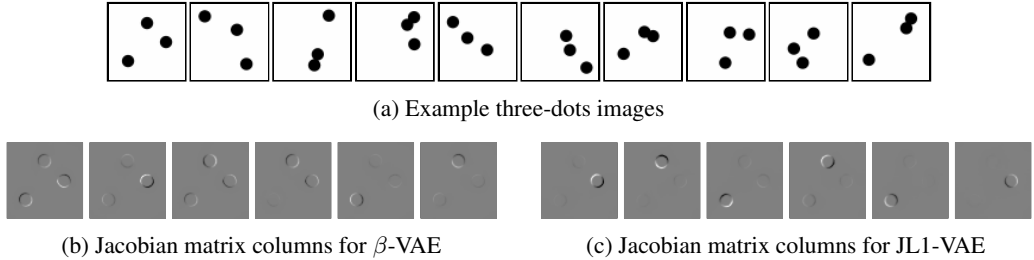
%
\centering%
\begin{subfigure}[]{0.98\textwidth}%
\centering%
\begin{tikzpicture}
    \foreach \x in {0,...,9} {
      \node[draw=black, inner sep=0, line width=0.6mm] at (\x * \boxscale,0)
      {\includegraphics[width=\imscale cm]{threeDotsJacobians/Fig2-ExampleThreeDots\x}};
  }
\end{tikzpicture}%
\caption{Example three-dots images}%
\end{subfigure}%
\medskip

\noindent 
\begin{subfigure}[]{0.48\textwidth}%
\centering%
\begin{tikzpicture}
    \foreach \x in {0,...,5} {
      \node[draw=black, inner sep=0] at (\x * \boxscale,0)
      {\includegraphics[width=\imscale cm]{threeDotsJacobians/Fig2-JacGamma0_0000Latent\x}};
  }
\end{tikzpicture}%
\caption{Jacobian matrix columns for $\beta$-VAE}%
\end{subfigure}%
\hspace{0.4cm}%
\begin{subfigure}[]{0.48\textwidth}%
\centering%
\begin{tikzpicture}
    \foreach \x in {0,...,5} {
      \node[draw=black, inner sep=0] at (\x * \boxscale,0)
      {\includegraphics[width=\imscale cm]{threeDotsJacobians/Fig2-JacGamma0_1000Latent\x}};
    }
\end{tikzpicture}%
\caption{Jacobian matrix columns for JL1-VAE}%
\end{subfigure}%
\caption{Qualitative results for the three-dots dataset. 
  We show six Jacobian matrix columns for $\beta$-VAE ($\beta=4$) and JL1-VAE ($\beta=4$, $\gamma=0.1$)
  evaluated for the leftmost example image.
  }%
\label{fig:jacobianVisualizationThreeDots}%
\end{figure}%
\def\boxscale{2.6}%
\def\imscale{2.5}%
\begin{figure}[t]%
\centering%
\begin{subfigure}[]{0.30\textwidth}%
\centering%
\begin{tikzpicture}
      \node[draw=black, inner sep=0] at (0,0)
      {\includegraphics[width=\imscale
      cm]{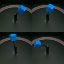}};
\end{tikzpicture}
\caption{MPI3D-Multi image}%
\end{subfigure}%
\begin{subfigure}[]{0.30\textwidth}%
\centering%
\begin{tikzpicture}
      \node[draw=black, inner sep=0] at (0,0)
      {\includegraphics[width=\imscale
      cm]{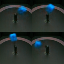}};
\end{tikzpicture}
\caption{$\beta$-VAE reconstruction}%
\end{subfigure}%
\begin{subfigure}[]{0.30\textwidth}%
\centering%
\begin{tikzpicture}
      \node[draw=black, inner sep=0] at (0,0)
      {\includegraphics[width=\imscale
      cm]{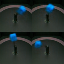}};
\end{tikzpicture}
\caption{JL1-VAE reconstruction}%
\end{subfigure}%
\medskip

\noindent 
\def\boxscale{1.2}%
\def\imscale{1.1}%
\begin{subfigure}[]{0.48\textwidth}%
\setcounter{imcnt}{0}%
\centering%
\begin{tikzpicture}
    \foreach \y in {0,...,1} {
    \foreach \x in {0,...,4} {
      \node[draw=black, inner sep=0] at (\x * \boxscale,-\y * \boxscale)
      {\includegraphics[width=\imscale cm]{Mpi3dMultiJacobians/Mpi3dMulti-JacGamma0_0000Latent\theimcnt}};
        \stepcounter{imcnt}
  }
}
\end{tikzpicture}%
\caption{Jacobian matrix columns for a $\beta$-VAE}%
\end{subfigure}%
\hspace{0.4cm}%
\begin{subfigure}[]{0.48\textwidth}%
\centering%
\setcounter{imcnt}{0}%
\begin{tikzpicture}
    \foreach \y in {0,...,1} {
    \foreach \x in {0,...,4} {
      \node[draw=black, inner sep=0] at (\x * \boxscale,-\y * \boxscale)
      {\includegraphics[width=\imscale cm]{Mpi3dMultiJacobians/Mpi3dMulti-JacGamma0_0100Latent\theimcnt}};
        \stepcounter{imcnt}
    }
  }
\end{tikzpicture}%
\caption{Jacobian matrix columns for a JL1-VAE}%
\end{subfigure}%
\caption{
  Qualitative results for MPI3D-Multi.
  JL1-VAE shows stronger pressure to locally disentangle 
  individual robot
  motions. 
  Both models used $\beta=0.01$. For JL1-VAE,
  $\gamma=0.01$. 
  }%
\label{fig:jacobianVisualizationMPI3DMulti}%
\end{figure}%

\subsection{Quantitative Results}
We generate local disentanglement scores for the models trained on the three-dots images.

\begin{figure}[ht]%
\centering%
\begin{subfigure}[]{0.47\textwidth}%
\begin{tikzpicture}
    \node
    {\includegraphics[width=0.95\textwidth]{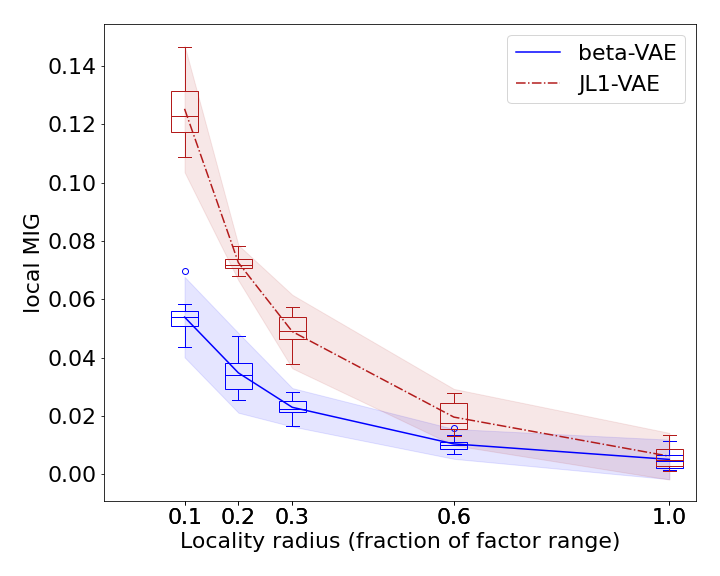}};
\end{tikzpicture}%
\caption{Local MIG scores}%
\end{subfigure}%
\begin{subfigure}[]{0.47\textwidth}%
\begin{tikzpicture}
    \node
    {\includegraphics[width=0.95\textwidth]{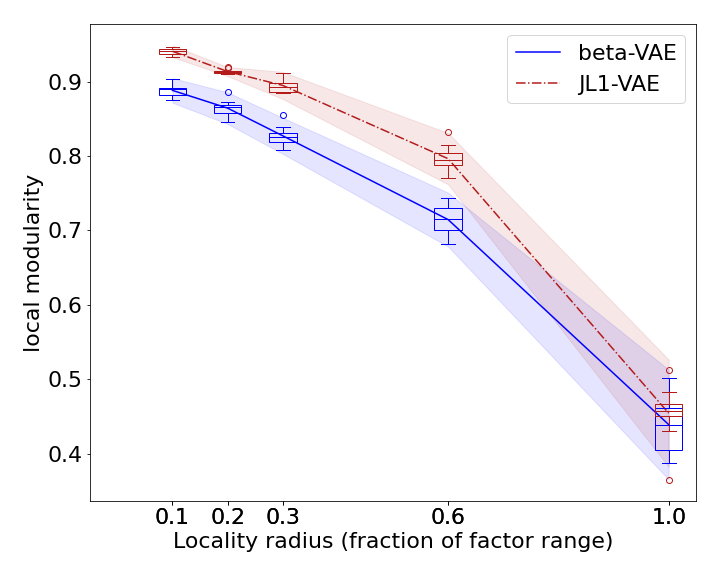}};
\end{tikzpicture}%
\caption{Local modularity scores}%
\end{subfigure}
\caption{Local disentanglement scores varying the locality
  parameter $\rho$. Ten JL1-VAE and $\beta$-VAE models were trained on the three-dots dataset with $\beta=4$ and, for JL1-VAE, $\gamma=0.1$.}
\label{fig:localMetricVaryingRho}
\end{figure}

To observe the effect of the $\rho$ parameter of our local disentanglement
metrics, in 
Figure~\ref{fig:localMetricVaryingRho}.
we plot the varying local disentanglement scores as we change the $\rho$
parameter for our JL1-VAE ($\beta=4.0$ and $\gamma=0.1$) and a standard $\beta$-VAE ($\beta=4.0$).
The hyperparameter for $\beta$ was chosen near the middle of the range used in~\cite{locatello2019challenging}. 
We also tried other hyperparameter values and saw similar results.
Too large a $\gamma$ can lead to model collapse, so we chose a small enough $\gamma$ to avoid that collapse but otherwise large enough to start to see reconstruction performance degradation, so we knew that its regularization was affecting model training.

We note quantitatively that JL1-VAE attains higher local disentanglement scores
compared to $\beta$-VAEs,
which is especially true as we look at more localized samples of data,
corresponding to a smaller
$\rho$ parameter.
For $\rho = 0.1$ we see significantly higher local disentanglement scores
for JL1-VAE compared to $\beta$-VAE
($p<0.001$ for T-test), but 
for $\rho = 1$, testing for global disentanglement, we see 
indistinguishable disentanglement scores between the two ($p > 0.05$ for
T-Test).
Our JL1-VAE is able to \emph{locally} disentangle factors of
variation for the three-dots dataset, but does not globally disentangle
factors of variation.

Fixing the $\rho$ parameter to $0.1$, we also compute the local MIG and local
modularity scores for six different comparative methods, namely $\beta$-VAE, FactorVAE, DIP-VAE-I, DIP-VAE-II, $\beta$-TCVAE, and AnnealedVAE, using the implementations of \texttt{disentanglement\_lib} with our convolutional architecture. A description of each of these models can be found in~\cite{locatello2019challenging}.
We trained ten iterations of those models
with different random seeds
using hyperparameters chosen near the middle of the suggested ranges in that
work.
That included training ten new $\beta$-VAE models with new random seeds.
All models were trained with ten latent dimensions.

\begin{figure}[ht]%
\centering%
\begin{subfigure}[]{0.49\textwidth}%
\begin{tikzpicture}
    \node {\includegraphics[width=0.95\textwidth]{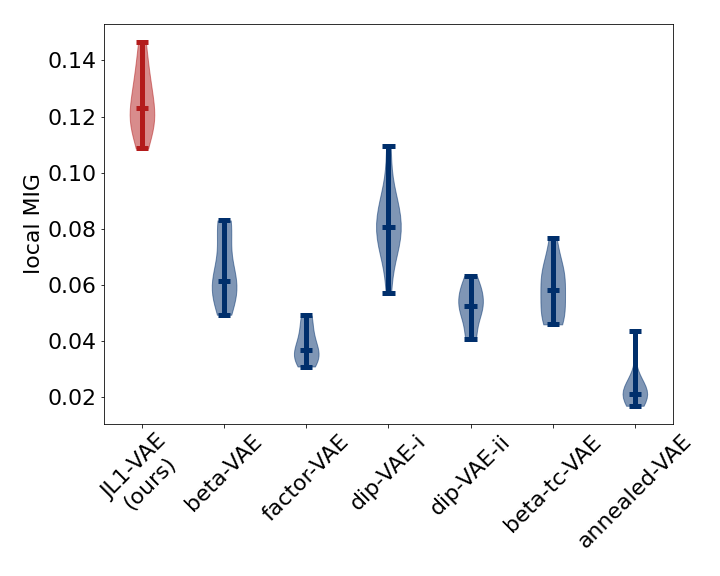}};
\end{tikzpicture}%
\caption{Local MIG scores}%
\end{subfigure}%
\begin{subfigure}[]{0.49\textwidth}%
\begin{tikzpicture}
    \node {\includegraphics[width=0.95\textwidth]{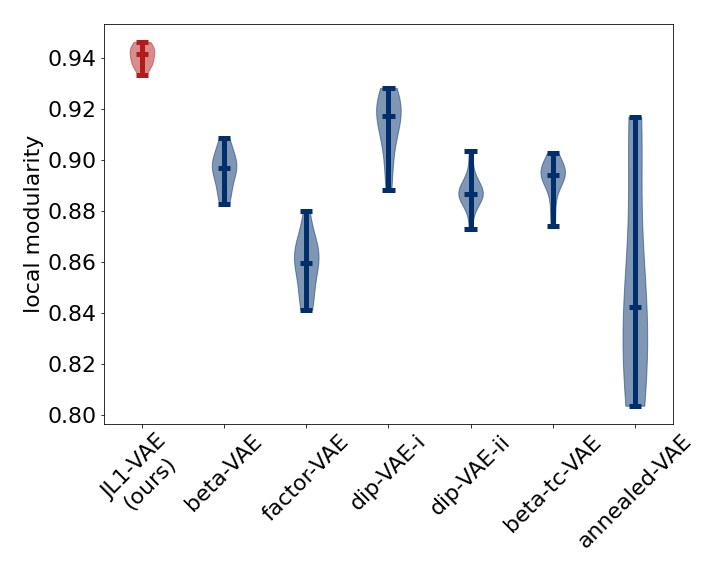}};
\end{tikzpicture}%
\caption{Local modularity scores}%
\end{subfigure}
\caption{Local disentanglement scores for JL1-VAE models and baseline
  implementations from~\cite{locatello2019challenging}. The baseline
  implementations use default hyperparameters from that paper, choosing values
  near the middle when a range of hyperparameters are listed. Each model is
  run ten times with new random seeds. Local disentanglement
  is calculated using $\rho=0.1$ with 20 different local samples.}
\label{fig:localMetricComparedToStandardMethods}
\end{figure}

Local MIG and local modularity scores are shown in
Figure~\ref{fig:localMetricComparedToStandardMethods}.
We see a range of disentanglement scores due to the random
seeds used to generate our models (ten models for each learning algorithm).
Additionally, our local MIG and modularity metrics have some additional
stochasticity due to the
randomness in sampling 20 local samples of
10,000 points during calculation of those metrics.
Nevertheless, we observe significantly higher
disentanglement scores ($p<0.001$ for T-test) for our JL1-VAE compared to every
baseline method.

\section{Discussion}
\label{sec:Discussion}
In this work, we presented JL1-VAE, a VAE augmented with an $L_1$  regularization to the Jacobian to improve local disentanglement.
We extended the MIG and modularity disentanglement metrics to generate metrics that can measure local disentanglement.
We evaluated our model on natural images, simulated images of dots, and tiled images of a real robot, and showed qualitatively and quantitatively 
that our method can improve local disentanglement in the generated representation.

Our added $L_1$ regularization to the Jacobian of the generator function
is motivated by 
the use of $L_1$ regularization to prefer certain orientations in ICA and sparse coding,
by a desire to relate each latent direction to sparser pixels (more similar to localized receptive fields),
and by 
the implicit $L_2$ regularization already present in $\beta$-VAEs.
We show that this $L_1$ regularization term can encourage latent axes to locally align with ground-truth factors of variation.
While this approach shows promise for local alignment, it does not address global alignment issues.
That is, in one part of the dataset, the learned representation may assign a latent variable $e_1$ to follow a certain latent factor of variation, and in a different part of the dataset it might be a different latent variable $e_2$ that follows that latent factor of variation.

Regarding ``no free lunch'' theorems that show unsupervised disentanglement is impossible without inductive biases~\cite{locatello2019challenging}, we note that $L_1$ regularization of the generative Jacobian generates an inductive bias. The inductive bias encourages small axis-aligned perturbations of the latent space to result in sparse changes to the image space, whether that be expressed as localized receptive fields which act only on small regions of the image space, as seen in Figure~\ref{fig:naturalImageJacobianColumns}, or motions of only a single dot or robot, as seen in Figures~\ref{fig:jacobianVisualizationThreeDots} and \ref{fig:jacobianVisualizationMPI3DMulti}.
Cases where this inductive bias might not add value would include, for example, single robot images from MPI3D, where the primary ground-truth factors of local variation affect the same object within the same image patch, or images where one factor of variation includes global lighting changes.

This work has only applied this learning method to image data, and we would like to apply this approach to multimodal data gathered from a real robotic system to understand broader applications.

Finally, this approach computes the full Jacobian of the generator during training, leading to training times that scale linearly with the number of latent dimensions. Future work should look to sampling-based methods to approximate the $L_1$ cost by only computing a subset of the Jacobian values in order to speed up training.


\bibliography{neurips_2021}

\begin{thebibliography}{38}
\expandafter\ifx\csname natexlab\endcsname\relax\def\natexlab#1{#1}\fi
\expandafter\ifx\csname url\endcsname\relax
  \def\url#1{{\tt #1}}\fi

\bibitem[Kingma and Welling(2014)]{kingma2014autoencoding}
Diederik~P Kingma and Max Welling.
\newblock Auto-encoding variational bayes, 2014, 1312.6114.

\bibitem[Higgins et~al.(2017)Higgins, Matthey, Pal, Burgess, Glorot, Botvinick,
  Mohamed, and Lerchner]{Higgins2017}
Irina Higgins, Loic Matthey, Arka Pal, Christopher Burgess, Xavier Glorot,
  Matthew Botvinick, Shakir Mohamed, and Alexander Lerchner.
\newblock {$\beta$-VAE: Learning basic visual concepts with a constrained
  variational framework}.
\newblock In {\em 5th International Conference on Learning Representations,
  ICLR 2017 - Conference Track Proceedings}, 2017.

\bibitem[Rolinek et~al.(2019)Rolinek, Zietlow, and Martius]{Rolinek2019}
Michal Rolinek, Dominik Zietlow, and Georg Martius.
\newblock {Variational autoencoders pursue pca directions (by accident)}.
\newblock In {\em Proceedings of the IEEE Computer Society Conference on
  Computer Vision and Pattern Recognition}, volume 2019-June, pages
  12398--12407, 2019.
\newblock ISBN 9781728132938.

\bibitem[Kumar and Poole(2020)]{pmlr-v119-kumar20d}
Abhishek Kumar and Ben Poole.
\newblock On implicit regularization in $\beta$-{VAE}s.
\newblock In Hal~Daumé III and Aarti Singh, editors, {\em Proceedings of the
  37th International Conference on Machine Learning}, volume 119 of {\em
  Proceedings of Machine Learning Research}, pages 5480--5490. PMLR, 13--18 Jul
  2020.
\newblock URL \url{http://proceedings.mlr.press/v119/kumar20d.html}.

\bibitem[Hyv{\"a}rinen and Oja(2000)]{hyvarinen2000independent}
Aapo Hyv{\"a}rinen and Erkki Oja.
\newblock Independent component analysis: algorithms and applications.
\newblock {\em Neural networks}, 13\penalty0 (4-5):\penalty0 411--430, 2000.

\bibitem[Olshausen(1996)]{OlshausenMIT1996}
Bruno~A Olshausen.
\newblock {Learning linear, sparse, factorial codes}.
\newblock Technical Report A.I. Memo No. 1580, MIT, September 1996.
\newblock URL \url{http://dspace.mit.edu/handle/1721.1/7184}.

\bibitem[Ols\-hausen and Field(1996)]{Olshausen1996}
Bruno~A. Ols\-hausen and David~J. Field.
\newblock {Emergence of simple-cell receptive field properties by learning a
  sparse code for natural images}.
\newblock {\em Nature}, 381\penalty0 (6583):\penalty0 607--609, 1996.
\newblock ISSN 00280836.

\bibitem[Jolliffe et~al.(2003)Jolliffe, Trendafilov, and Uddin]{Jolliffe2003}
Ian~T Jolliffe, Nickolay~T Trendafilov, and Mudassir Uddin.
\newblock {A Modified Principal Component Technique Based on the LASSO}.
\newblock {\em Journal of Computational and Graphical Statistics}, 12\penalty0
  (3):\penalty0 531--547, 2003.
\newblock ISSN 10618600.

\bibitem[Zou et~al.(2006)Zou, Hastie, and Tibshirani]{Zou2006}
Hui Zou, Trevor Hastie, and Robert Tibshirani.
\newblock {Sparse principal component analysis}.
\newblock {\em Journal of Computational and Graphical Statistics}, 15\penalty0
  (2):\penalty0 265--286, 2006.
\newblock ISSN 10618600.

\bibitem[Seung and Lee(2000)]{manifoldPerceptionLee}
H.~Sebastian Seung and Daniel~D. Lee.
\newblock The manifold ways of perception.
\newblock {\em Science}, 290\penalty0 (5500):\penalty0 2268--2269, 2000,
  https://www.science.org/doi/pdf/10.1126/science.290.5500.2268.
\newblock URL
  \url{https://www.science.org/doi/abs/10.1126/science.290.5500.2268}.

\bibitem[Locatello et~al.(2019)Locatello, Bauer, Lucic, Raetsch, Gelly,
  Sch{\"o}lkopf, and Bachem]{locatello2019challenging}
Francesco Locatello, Stefan Bauer, Mario Lucic, Gunnar Raetsch, Sylvain Gelly,
  Bernhard Sch{\"o}lkopf, and Olivier Bachem.
\newblock Challenging common assumptions in the unsupervised learning of
  disentangled representations.
\newblock In {\em International Conference on Machine Learning}, pages
  4114--4124, 2019.

\bibitem[Mathieu et~al.(2019)Mathieu, Rainforth, Siddharth, and
  Teh]{Mathieu2019}
Emile Mathieu, Tom Rainforth, N~Siddharth, and Yee~Whye Teh.
\newblock {Disentangling disentanglement in variational autoencoders}.
\newblock In {\em 36th International Conference on Machine Learning, ICML
  2019}, volume 2019-June, pages 7744--7754, 2019, 1812.02833.
\newblock ISBN 9781510886988.
\newblock URL \url{http://github.com/iffsid/disentangling-disentanglement.}

\bibitem[Makhzani and Frey(2014)]{Makhzani2014}
Alireza Makhzani and Brendan Frey.
\newblock {k-Sparse autoencoders}.
\newblock In {\em 2nd International Conference on Learning Representations,
  ICLR 2014 - Conference Track Proceedings}, 2014, 1312.5663.

\bibitem[Ng(2011)]{SparseAutoEncoderNg}
Andrew Ng.
\newblock {CS294A Lecture Notes Sparse Autoencoder}.
\newblock {\em Cs294a}, pages 1--19, 2011, arXiv:1506.03733v1.
\newblock ISSN 19326203.
\newblock URL \url{http://www.stanford.edu/class/cs294a/}.

\bibitem[Jiang and de~la Iglesia(2021)]{jiang2021improved}
Linxing~Preston Jiang and Luciano de~la Iglesia.
\newblock Improved training of sparse coding variational autoencoder via weight
  normalization.
\newblock {\em CoRR}, abs/2101.09453, 2021, 2101.09453.
\newblock URL \url{https://arxiv.org/abs/2101.09453}.

\bibitem[Louizos et~al.(2018)Louizos, Welling, and Kingma]{Louizos2018}
Christos Louizos, Max Welling, and Diederik~P. Kingma.
\newblock {Learning sparse neural networks through L0 regularization}.
\newblock In {\em 6th International Conference on Learning Representations,
  ICLR 2018 - Conference Track Proceedings}, 2018, 1712.01312.
\newblock ISBN 1712.01312v2.

\bibitem[Denil et~al.(2013)Denil, Shakibi, Dinh, Ranzato, and {De
  Freitas}]{Denil2013}
Misha Denil, Babak Shakibi, Laurent Dinh, Marc'aurelio Ranzato, and Nando {De
  Freitas}.
\newblock {Predicting parameters in deep learning}.
\newblock In {\em Advances in Neural Information Processing Systems}, 2013,
  1306.0543.

\bibitem[Kim and Mnih(2018)]{Kim2018}
Hyunjik Kim and Andriy Mnih.
\newblock {Disentangling by factorising}.
\newblock In {\em 35th International Conference on Machine Learning, ICML
  2018}, volume~6, pages 4153--4171, 2018, 1802.05983.
\newblock ISBN 9781510867963.

\bibitem[Locatello et~al.(2020)Locatello, Poole, Raetsch, Sch{\"o}lkopf,
  Bachem, and Tschannen]{pmlr-v119-locatello20a}
Francesco Locatello, Ben Poole, Gunnar Raetsch, Bernhard Sch{\"o}lkopf, Olivier
  Bachem, and Michael Tschannen.
\newblock Weakly-supervised disentanglement without compromises.
\newblock In Hal~Daumé III and Aarti Singh, editors, {\em Proceedings of the
  37th International Conference on Machine Learning}, volume 119 of {\em
  Proceedings of Machine Learning Research}, pages 6348--6359. PMLR, 13--18 Jul
  2020.
\newblock URL \url{http://proceedings.mlr.press/v119/locatello20a.html}.

\bibitem[Stuehmer et~al.(2020)Stuehmer, Turner, and
  Nowozin]{pmlr-v108-stuehmer20a}
Jan Stuehmer, Richard Turner, and Sebastian Nowozin.
\newblock Independent subspace analysis for unsupervised learning of
  disentangled representations.
\newblock In Silvia Chiappa and Roberto Calandra, editors, {\em Proceedings of
  the Twenty Third International Conference on Artificial Intelligence and
  Statistics}, volume 108 of {\em Proceedings of Machine Learning Research},
  pages 1200--1210. PMLR, 26--28 Aug 2020.
\newblock URL \url{http://proceedings.mlr.press/v108/stuehmer20a.html}.

\bibitem[Khemakhem et~al.(2020)Khemakhem, Kingma, Monti, and
  Hyvarinen]{pmlr-v108-khemakhem20a}
Ilyes Khemakhem, Diederik Kingma, Ricardo Monti, and Aapo Hyvarinen.
\newblock Variational autoencoders and nonlinear ica: A unifying framework.
\newblock In Silvia Chiappa and Roberto Calandra, editors, {\em Proceedings of
  the Twenty Third International Conference on Artificial Intelligence and
  Statistics}, volume 108 of {\em Proceedings of Machine Learning Research},
  pages 2207--2217. PMLR, 26--28 Aug 2020.
\newblock URL \url{http://proceedings.mlr.press/v108/khemakhem20a.html}.

\bibitem[Greff et~al.(2019)Greff, Kaufman, Kabra, Watters, Burgess, Zoran,
  Matthey, Botvinick, and Lerchner]{pmlr-v97-greff19a}
Klaus Greff, Rapha{\"e}l~Lopez Kaufman, Rishabh Kabra, Nick Watters,
  Christopher Burgess, Daniel Zoran, Loic Matthey, Matthew Botvinick, and
  Alexander Lerchner.
\newblock Multi-object representation learning with iterative variational
  inference.
\newblock In Kamalika Chaudhuri and Ruslan Salakhutdinov, editors, {\em
  Proceedings of the 36th International Conference on Machine Learning},
  volume~97 of {\em Proceedings of Machine Learning Research}, pages
  2424--2433. PMLR, 09--15 Jun 2019.
\newblock URL \url{http://proceedings.mlr.press/v97/greff19a.html}.

\bibitem[Watters et~al.(2019)Watters, Matthey, Burgess, and
  Lerchner]{watters2019spatial}
Nicholas Watters, Loic Matthey, Christopher~P. Burgess, and Alexander Lerchner.
\newblock Spatial broadcast decoder: A simple architecture for learning
  disentangled representations in vaes, 2019, 1901.07017.

\bibitem[Ainsworth et~al.(2018)Ainsworth, Foti, Lee, and Fox]{Ainsworth2018}
Samuel~K. Ainsworth, Nicholas~J. Foti, Adrian K.~C. Lee, and Emily~B. Fox.
\newblock oi-{VAE}: Output interpretable {VAE}s for nonlinear group factor
  analysis.
\newblock In Jennifer Dy and Andreas Krause, editors, {\em Proceedings of the
  35th International Conference on Machine Learning}, volume~80 of {\em
  Proceedings of Machine Learning Research}, pages 119--128. PMLR, 10--15 Jul
  2018.
\newblock URL \url{http://proceedings.mlr.press/v80/ainsworth18a.html}.

\bibitem[Khan et~al.(2021)Khan, Anwaar, and Kleinsteuber]{khan2021}
Rayyan~Ahmad Khan, Muhammad~Umer Anwaar, and Martin Kleinsteuber.
\newblock Epitomic variational graph autoencoder.
\newblock In {\em 2020 25th International Conference on Pattern Recognition
  (ICPR)}, pages 7203--7210, 2021.

\bibitem[Tomczak and Welling(2018)]{Tomczak2017}
Jakub~M. Tomczak and Max Welling.
\newblock {VAE with a vampprior}.
\newblock In {\em International Conference on Artificial Intelligence and
  Statistics, AISTATS 2018}, pages 1214--1223, may 2018, 1705.07120.
\newblock URL \url{http://arxiv.org/abs/1705.07120}.

\bibitem[Bauer and Mnih(2020)]{Bauer2018}
Matthias Bauer and Andriy Mnih.
\newblock {Resampled priors for variational autoencoders}.
\newblock In {\em AISTATS 2019 - 22nd International Conference on Artificial
  Intelligence and Statistics}, oct 2020, 1810.11428.
\newblock URL \url{http://arxiv.org/abs/1810.11428}.

\bibitem[Davidson et~al.(2018)Davidson, Falorsi, {De Cao}, Kipf, and
  Tomczak]{Davidson2018}
Tim~R. Davidson, Luca Falorsi, Nicola {De Cao}, Thomas Kipf, and Jakub~M.
  Tomczak.
\newblock {Hyperspherical variational auto-encoders}.
\newblock In {\em 34th Conference on Uncertainty in Artificial Intelligence
  2018, UAI 2018}, volume~2, pages 856--865, apr 2018, 1804.00891.
\newblock ISBN 9781510871601.
\newblock URL \url{http://arxiv.org/abs/1804.00891}.

\bibitem[{Perez Rey} et~al.(2020){Perez Rey}, Menkovski, and Portegies]{Reyb}
Luis~A. {Perez Rey}, Vlado Menkovski, and Jim Portegies.
\newblock {Diffusion variational autoencoders}.
\newblock In {\em IJCAI International Joint Conference on Artificial
  Intelligence}, volume 2021-Janua, pages 2704--2710, 2020, 1901.08991.
\newblock ISBN 9780999241165.

\bibitem[Rifai et~al.(2011)Rifai, Vincent, Muller, Glorot, and
  Bengio]{Rifai2011}
Salah Rifai, Pascal Vincent, Xavier Muller, Xavier Glorot, and Yoshua Bengio.
\newblock {Contractive auto-encoders: Explicit invariance during feature
  extraction}.
\newblock In {\em Proceedings of the 28th International Conference on Machine
  Learning, ICML 2011}, pages 833--840, 2011.
\newblock ISBN 9781450306195.

\bibitem[Varga et~al.(2018)Varga, Csisz{\'{a}}rik, and Zombori]{Varga2018}
D{\'{a}}niel Varga, Adri{\'{a}}n Csisz{\'{a}}rik, and Zsolt Zombori.
\newblock {Gradient Regularization Improves Accuracy of Discriminative Models}.
\newblock {\em Schedae Informaticae}, 27:\penalty0 31--45, 2018, 1712.09936.
\newblock ISSN 20838476.

\bibitem[Hoffman et~al.(2019)Hoffman, Roberts, and Yaida]{hoffman2019robust}
Judy Hoffman, Daniel~A. Roberts, and Sho Yaida.
\newblock Robust learning with jacobian regularization, 2019, 1908.02729.

\bibitem[Chen et~al.(2020)Chen, Klushyn, Ferroni, Bayer, and van~der
  Smagt]{Chen2020}
Nutan Chen, Alexej Klushyn, Francesco Ferroni, Justin Bayer, and Patrick
  van~der Smagt.
\newblock {Learning flat latent manifolds with VAEs}.
\newblock In {\em 37th International Conference on Machine Learning, ICML
  2020}, volume PartF16814, pages 1565--1574, 2020, 2002.04881.
\newblock ISBN 9781713821120.

\bibitem[Zhao et~al.(2018)Zhao, Ren, Yuan, Song, Goodman, and Ermon]{Zhao2018}
Shengjia Zhao, Hongyu Ren, Arianna Yuan, Jiaming Song, Noah~D. Goodman, and
  Stefano Ermon.
\newblock Bias and generalization in deep generative models: An empirical
  study.
\newblock {\em CoRR}, abs/1811.03259, 2018, 1811.03259.
\newblock URL \url{http://arxiv.org/abs/1811.03259}.

\bibitem[Gondal et~al.(2019)Gondal, Wuthrich, Miladinovic, Locatello, Breidt,
  Volchkov, Akpo, Bachem, Sch\"{o}lkopf, and Bauer]{Gondal2019}
Muhammad~Waleed Gondal, Manuel Wuthrich, Djordje Miladinovic, Francesco
  Locatello, Martin Breidt, Valentin Volchkov, Joel Akpo, Olivier Bachem,
  Bernhard Sch\"{o}lkopf, and Stefan Bauer.
\newblock On the transfer of inductive bias from simulation to the real world:
  a new disentanglement dataset.
\newblock In H.~Wallach, H.~Larochelle, A.~Beygelzimer, F.~d'~Alch\'{e}-Buc,
  E.~Fox, and R.~Garnett, editors, {\em Advances in Neural Information
  Processing Systems}, volume~32. Curran Associates, Inc., 2019.
\newblock URL
  \url{https://proceedings.neurips.cc/paper/2019/file/d97d404b6119214e4a7018391195240a-Paper.pdf}.

\bibitem[Chen et~al.(2018)Chen, Li, Grosse, and Duvenaud]{chen2018isolating}
Ricky~TQ Chen, Xuechen Li, Roger Grosse, and David Duvenaud.
\newblock Isolating sources of disentanglement in vaes.
\newblock In {\em Proceedings of the 32nd International Conference on Neural
  Information Processing Systems}, pages 2615--2625, 2018.

\bibitem[Ridgeway and Mozer(2018)]{ridgewayNeurIPS2018}
Karl Ridgeway and Michael~C. Mozer.
\newblock Learning deep disentangled embeddings with the f-statistic loss.
\newblock In {\em Proceedings of the 32nd International Conference on Neural
  Information Processing Systems}, NIPS'18, page 185–194, Red Hook, NY, USA,
  2018. Curran Associates Inc.

\bibitem[Pedregosa et~al.(2011)Pedregosa, Varoquaux, Gramfort, Michel, Thirion,
  Grisel, Blondel, Prettenhofer, Weiss, Dubourg, Vanderplas, Passos,
  Cournapeau, Brucher, Perrot, and Duchesnay]{scikit-learn}
F.~Pedregosa, G.~Varoquaux, A.~Gramfort, V.~Michel, B.~Thirion, O.~Grisel,
  M.~Blondel, P.~Prettenhofer, R.~Weiss, V.~Dubourg, J.~Vanderplas, A.~Passos,
  D.~Cournapeau, M.~Brucher, M.~Perrot, and E.~Duchesnay.
\newblock Scikit-learn: Machine learning in {P}ython.
\newblock {\em Journal of Machine Learning Research}, 12:\penalty0 2825--2830,
  2011.

\end{thebibliography}

\appendix
\section{Neural Network Architecture}
We use a convolutional neural network architecture for our models. Our code can be found in our open source repository\footnote{https://github.com/travers-rhodes/jlonevae}

For the auto-encoders used on 64$\times$64-pixel images, we mimic the
architecture presented in~\cite{locatello2019challenging}.
We use 4$\times$4 kernels for all convolutional layers with a stride of
2.
We use a ReLU between all layers, with a final sigmoidal layer on the
reconstruction architecture and a Bernoulli
loss.
In Tables~\ref{tab:convEmbArchitecture} and~\ref{tab:convEmbArchitectureTiny}, ``Conv2d'' refers to a convolutional layer,
``FC'' refers to a fully connected layer, ``ConvT2d'' refers to convolutional
transpose, and the ``($\times$ 2)'' in the
embedding architecture refers to the separate mean and log variance
heads on the shared architecture.

\begin{table}[ht]
\caption{Embedding and reconstruction architectures for 64$\times$64-pixel
  images}
\label{tab:bigArchitectureTable}
\medskip

\noindent 
\label{tab:convEmbArchitecture}
\centering
  \begin{tabular}{p{0.4\linewidth}p{0.4\linewidth}}
\toprule
  \textbf{Embedding}&
  \textbf{Reconstruction}\\
    \midrule
    Input:  64$\times$64, 1 or 3 channels &
Input: 10 values\\
  Conv2d: 32 channels &
FC: 256 channels \\
  Conv2d:  32 channels &
FC:  4$\times$4 image, 64 channels\\
  Conv2d: 64 channels &
ConvT2d:  64 channels\\
  Conv2d: 64 channels& 
ConvT2d: 32 channels\\
  FC: 256 channels&
ConvT2d: 32 channels\\
  FC ($\times$ 2): 10 values &
ConvT2d: 64$\times$64, 1 or 3 channels\\
\bottomrule
\end{tabular}
\end{table}

For the auto-encoders used on 16$\times$16-pixel images (the natural image
crops), we use 3$\times$3 kernels for all convolutional layers and a stride of 2
everywhere except for the last reconstruction layer, which has a stride of 1.
We use a ReLU between all layers, with a final sigmoidal layer on the
reconstruction architecture and a Bernoulli
loss.

\begin{table}[ht]
\caption{Embedding and reconstruction architectures for 16$\times$16-pixel
  images}
\medskip

\noindent 
\label{tab:convEmbArchitectureTiny}
\centering
  \begin{tabular}{p{0.4\linewidth}p{0.4\linewidth}}
\toprule
  \textbf{Embedding}&
  \textbf{Reconstruction}\\
    \midrule
    Input:  16$\times$16, 1 channel &
Input: 10 values\\
  Conv2d: 64 channels &
FC: 128 channels \\
  Conv2d:  128 channels &
FC:  4$\times$4 image, 64 channels\\
  FC: 128 channels&
ConvT2d: 64 channels\\
  FC ($\times$ 2): 10 values &
ConvT2d: 32 channels\\
    &
ConvT2d: stride 1, 64$\times$64, 1 channel\\
\bottomrule
\end{tabular}
\end{table}

For the loss, We estimate the full Jacobian matrix $J_g(z)$ using the finite difference method
along each latent dimension.
That is, for any given latent value $z$ at which we wish to compute the Jacobian matrix,
we generate a set of $k$ data points $z_i = z+\epsilon e_i$, where $\epsilon$ is a small fixed value and $e_i$ a unit vector in the $i$\textsuperscript{th} latent direction.
We then run the forward model on the batch of $z_i$ to generate $g(z_i)$ and estimate the $i$\textsuperscript{th} column of the Jacobian matrix as $(g(z_i) - g(z))/\epsilon$
This Jacobian matrix estimate is itself backward differentiable using standard backward differentiation, and can be directly used in our JL1-VAE loss.

\section{Three-dots Experiment Hyperparameters and Additional Results}

For the three-dots dataset, We discretize the possible x,y coordinates of the center of each dot to 64 different
values.
We note that the generative map is not injective, as the dots are identical, so
the same resulting image can be formed from multiple permutations of ground-truth factor values.
There are $64^6 \sim 68.7$ billion different possible input latent
combinations, from which we pre-generate a cache of 500,000 images on
which we train.
During evaluation, we generate new images at runtime based on the desired
ground-truth factors of variation. 

We train a $\beta$-VAE with $\beta = 4$ on a training dataset cache of 500,000
64$\times$64-pixel
images of three black dots on a white background, with $x$ and $y$ values for
the dot centers appearing independently at
one of 64 possible discrete locations, evenly spaced horizontally and vertically, across the image.
We embed the dataset into a latent space of 10 dimensions.
We train on 300,000 independently sampled batches of 64 images from the cache, giving a total of
19,200,000 image presentations to the neural network.
Additionally, we train our JL1-VAE with the same $\beta$ and model architecture on the same training dataset
with our added $L_1$ regularization weighted by a hyperparameter $\gamma = 0.1$.
The hyperparameter $\gamma$ was chosen as the largest tested for which the learning algorithm converged to give good reconstruction accuracy.
We use linear annealing for both the $\beta$ and $\gamma$ parameters, annealing
each from 0 to their final values over the first 100,000 batches.
We use the Adam optimizer with a learning rate of 0.0001.

For the baseline comparison models, we use the implementations of $\beta$-VAE,
FactorVAE, DIP-VAE-I, DIP-VAE-II, $\beta$-TCVAE, and AnnealedVAE
from~\cite{locatello2019challenging}, matching their hyperparameter choices.
For implementations that for which they provided a range of tested
hyperparameters, we chose near the middle of their range. Thus, for $\beta$-VAE
we used $\beta=4$; for Annealed VAE we use $c_{max}=25$, iteration threshold$=
100000$, and $\gamma=1000$; for Factor VAE we use
$\gamma=30$; for DIP-VAE-I we use $\lambda_{od}=5$ and $\lambda_d = 50$; for
DIP-VAE-II we use $\lambda_{od}=5$ and $\lambda_d = 5$; and for $\beta$-TCVAE we
use $\beta=4$.

We note that we modified the reference implementation provided with that work
from in order to have consistent 4$\times$4
kernels as shown in the architecture in Table~\ref{tab:bigArchitectureTable}. We include the modified implementation in our supplemental materials. 
The reference implementation of that architecture 
had unexplained 2$\times$2 convolutional kernels for two
layers.

We varied the random model initialization seed ten times and trained ten different
models for each algorithm type. Additionally, we ran a smaller
experiment randomizing both the model initialization
seed and using different initial seeds for data sampling as well, 
getting similar results to those presented in the paper.
The baseline implementation samples batches by epoch, shuffling after each
epoch, while our implementation pulls independent random batches at each
training set. Thus, the results for $\beta$-VAE in Figure~\ref{fig:localMetricVaryingRho} use
independently-sampled random batches, while the results for $\beta$-VAE
in~\ref{fig:localMetricComparedToStandardMethods}  use shuffling
after each epoch. This did not seem to affect results.

For the local metric calculations, we use the implementation provided
by~\cite{locatello2019challenging}. 
We sample 20 different local regions, pulling 10,000 points for each.
We use a histogram discretization with 5 bins for mutual information
calculations.

All ten Jacobian columns associated with
Figure~\ref{fig:jacobianVisualizationThreeDots} are shown in
Figure~\ref{fig:jacobianVisualizationThreeDotsAll}.
\def\boxscale{1.1}%
\def\imscale{1.0}%
\begin{figure}[t]
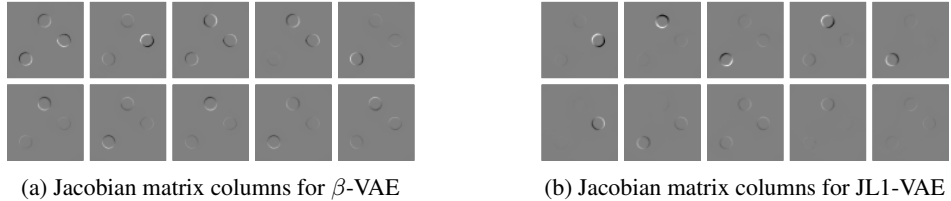
%
\centering%
\begin{subfigure}[]{0.48\textwidth}%
\centering%
\begin{tikzpicture}
  \setcounter{imcnt}{0}
    \foreach \y in {0,...,1} {
    \foreach \x in {0,...,4} {
      \node[draw=black, inner sep=0] at (\x * \boxscale,-\y * \boxscale)
      {\includegraphics[width=\imscale cm]{threeDotsJacobians/Fig2-JacGamma0_0000Latent\theimcnt}};
      \stepcounter{imcnt}
    }}
\end{tikzpicture}%
\caption{Jacobian matrix columns for $\beta$-VAE}%
\end{subfigure}%
\hspace{0.4cm}%
\begin{subfigure}[]{0.48\textwidth}%
\centering%
\begin{tikzpicture}
  \setcounter{imcnt}{0}
    \foreach \y in {0,...,1} {
    \foreach \x in {0,...,4} {
      \node[draw=black, inner sep=0] at (\x * \boxscale,-\y * \boxscale)
      {\includegraphics[width=\imscale cm]{threeDotsJacobians/Fig2-JacGamma0_1000Latent\theimcnt}};
      \stepcounter{imcnt}
    }}
\end{tikzpicture}%
\caption{Jacobian matrix columns for JL1-VAE}%
\end{subfigure}%
\caption{Qualitative results for three-dots. Both models used $\beta=4.0$. For JL1-VAE,
  $\gamma=0.1$. 
  All Jacobian matrix columns are shown.
  }%
\label{fig:jacobianVisualizationThreeDotsAll}%
\end{figure}%

\begin{figure}[ht]%
\centering%
\begin{subfigure}[]{0.45\textwidth}%
\begin{tikzpicture}
    \node
    {\includegraphics[width=0.95\textwidth]{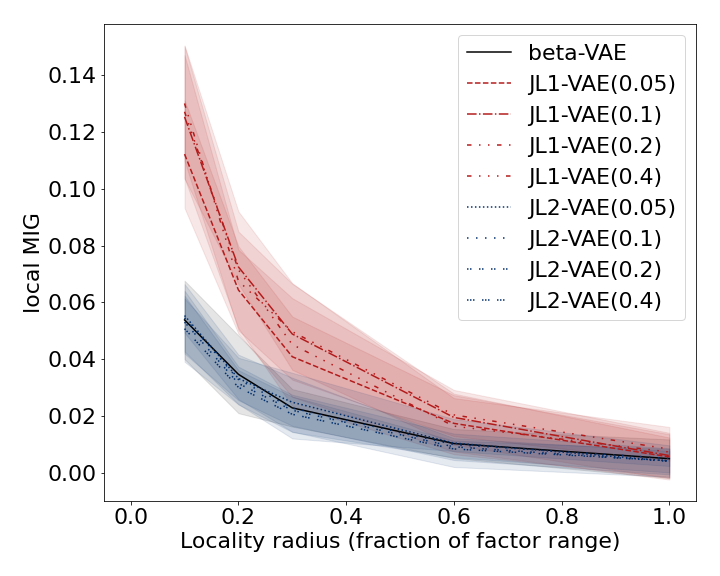}};
\end{tikzpicture}%
\caption{Local MIG scores}%
\end{subfigure}%
\begin{subfigure}[]{0.45\textwidth}%
\begin{tikzpicture}
    \node
    {\includegraphics[width=0.95\textwidth]{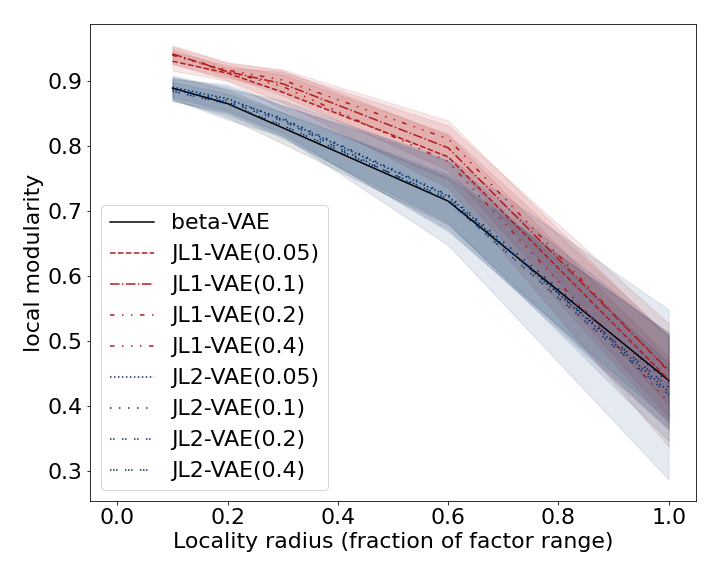}};
\end{tikzpicture}%
\caption{Local modularity scores}%
\end{subfigure}
\caption{Local disentanglement scores varying the locality
  parameter $\rho$ and the regularization factor $\gamma$ for JL1-VAE, JL2-VAE, and $\beta$-VAE. The regularization factor $\gamma$ is given in parentheses in the legend. Ten of each type of model were trained on the three-dots dataset with $\beta=4$. This figure best viewed in color.}
\label{fig:l2insteadofl1}
\end{figure}

Additionally, we validate that an $L_2$ loss does not have the same disentangling properties as our $L_1$ loss by replacing the $L_1$ loss with an $L_2$ loss and computing the local disentanglement metrics in Figure~\ref{fig:l2insteadofl1}. We label the JL1-VAE with $L_1$ replaced by $L_2$ a JL2-VAE. All the $L_2$ regularizations are roughly indistinguishable from the $\beta$-VAE result, while the JL1-VAE consistently outperforms for the full range of tested regularization values $\gamma$.

\section{Natural Image Experiment Hyperparameters and Additional Results}
We train a $\beta$-VAE with $\beta = 0.01$ on a dataset 100,000 16x16-pixel crops from grayscale natural scenes~\cite{Olshausen1996}, 
embedding the dataset into a latent space of 10 variables.
We train for 100,000 batches of 128 images, re-shuffling the images after each
epoch. 
We note that due to
epoch endings a few of the batches were incomplete, with fewer than 128 images.
Additionally, we train our JL1-VAE with the same model architecture  and $\beta$ on the same training dataset
with our added $L_1$ regularization cost weighted by a hyperparameter $\gamma = 0.01$.
The $\beta$ was chosen as large as possible that still avoided significant dimensionality collapse, and then the hyperparameter $\gamma$ was chosen as the largest tested for which the learning algorithm converged to give good reconstruction accuracy.
We use linear annealing for both the $\beta$ and $\gamma$ parameters, annealing
each from 0 to their final values over the first 50,000 batches.
We use the Adam optimizer with a learning rate of 0.001.

\begin{figure}[h]
  \centering
  \def \yimcropstart{100}%
  \def \ximcropstart{45}%
  \def \gammaval{00}%
  \setcounter{imcnt}{0}
  \begin{subfigure}[]{12cm}
    \centering
\begin{tikzpicture}
    \foreach \y in {0,...,0} {
      \foreach \x in {0,...,4} {
        \node[inner sep=0] at (\x, -\y)
         {\includegraphics[width=0.9cm]{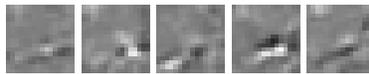}};
        \stepcounter{imcnt}
      }
    }
\end{tikzpicture}%

    \caption{Jacobian columns for a $\beta$-VAE with 5 latent dimensions} 
  \end{subfigure}
\medskip

\noindent 
  \setcounter{imcnt}{0}
  \begin{subfigure}[]{12cm}
    \centering
  \def \gammaval{10}%
  \begin{tikzpicture}
    \foreach \y in {0,...,0} {
      \foreach \x in {0,...,4} {
        \node[inner sep=0] at (\x, -\y)
         {\includegraphics[width=0.9cm]{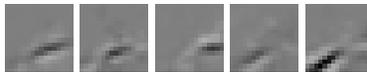}};
        \stepcounter{imcnt}
      }
    }
  \end{tikzpicture}
    \caption{Jacobian columns for a JL1-VAE with 5 latent dimensions} 
  \end{subfigure}
  \caption{Results for $\beta$-VAE and JL1-VAE using 5 latent dimensions
  (instead of the 10 shown in the main paper). Both are trained with
  $\beta=0.01$, and JL1-VAE trained with $\gamma=0.01$.
  }
  \label{fig:additionalNaturalImageLatents5}
\end{figure}

\begin{figure}[h]
  \centering
  \setcounter{imcnt}{0}
  \begin{subfigure}[]{12cm}
    \centering
  \begin{tikzpicture}
    \foreach \y in {0,...,9} {
      \foreach \x in {0,...,9} {
        \node[inner sep=0] at (\x * 1,-\y * 1) {\includegraphics[width=0.9cm]{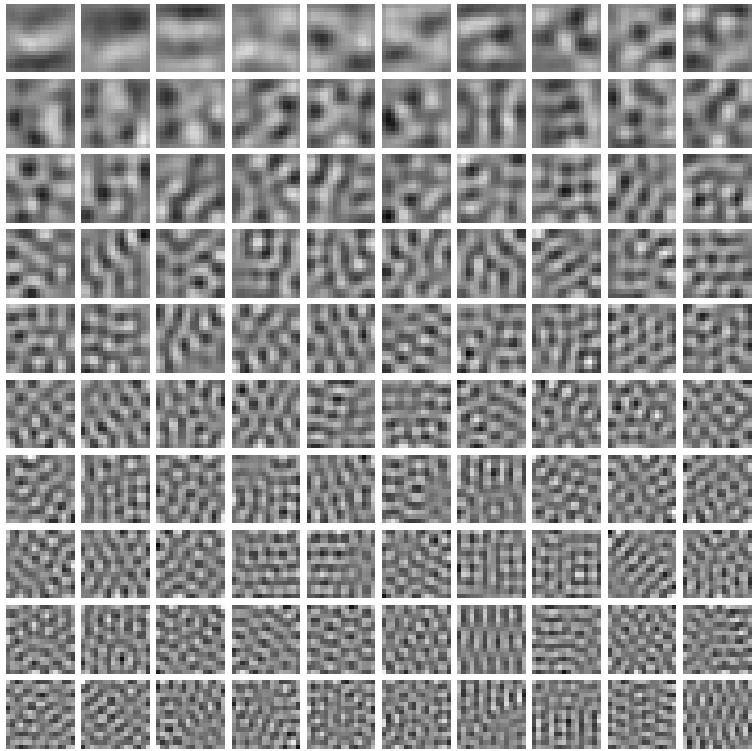}};
        \stepcounter{imcnt}
      }
    }
  \end{tikzpicture}
    \caption{100 latent PCA directions with the largest explained variance~\cite{scikit-learn}}
  \end{subfigure}
\medskip

\noindent 
  \setcounter{imcnt}{0}
  \begin{subfigure}[]{12cm}
    \centering
  \begin{tikzpicture}
    \foreach \y in {0,...,9} {
      \foreach \x in {0,...,9} {
        \node[inner sep=0] at (\x * 1,-\y * 1) {\includegraphics[width=0.9cm]{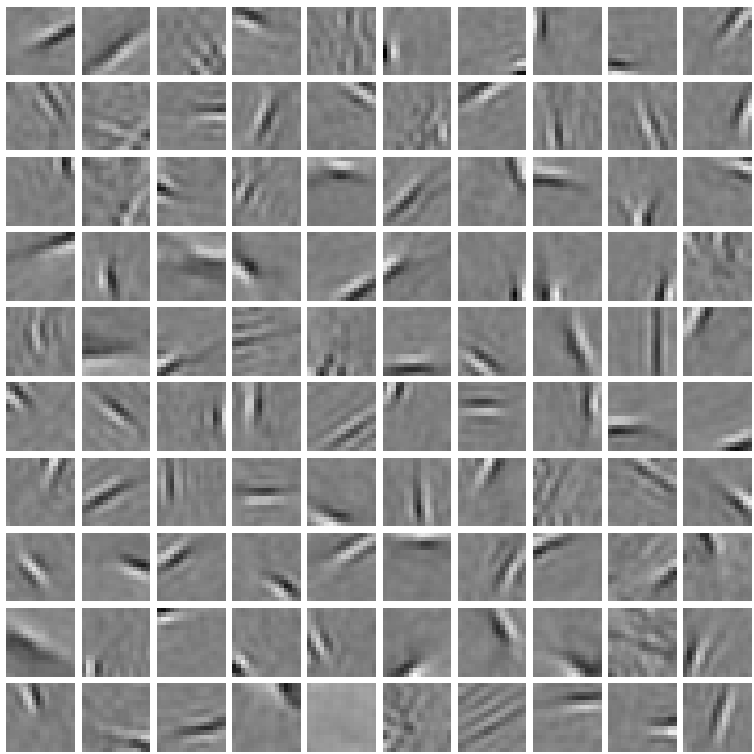}};
        \stepcounter{imcnt}
      }
    }
  \end{tikzpicture}
    \caption{100 latent ICA directions fit using FastICA\cite{scikit-learn,hyvarinen2000independent}} 
  \end{subfigure}
  \caption{Latent vectors for PCA and ICA trained on random 16x16 crops from natural images collected by~\cite{Olshausen1996}}
  \label{fig:linearModelNaturalImageFullJacobians}
\end{figure}

We show additional Jacobian column results, training on five latent dimensions in
Figure~\ref{fig:additionalNaturalImageLatents5},
and training on 25 latent dimensions in Figure~\ref{fig:additionalNaturalImageLatents25}.
We also show the top 100 PCA components and 100 trained ICA components in
Figure~\ref{fig:linearModelNaturalImageFullJacobians}.

\begin{figure}[h]
  \centering
  \def \yimcropstart{100}%
  \def \ximcropstart{45}%
  \def \gammaval{00}%
  \setcounter{imcnt}{0}
  \begin{subfigure}[]{12cm}
    \centering
\begin{tikzpicture}
    \foreach \y in {0,...,4} {
      \foreach \x in {0,...,4} {
        \node[inner sep=0] at (\x, -\y)
         {\includegraphics[width=0.9cm]{latentJacobianImages/naturalImageNorm/beta0_010_ica0_0\gammaval_lat25_im0_latind\theimcnt_x\ximcropstart_y\yimcropstart}};
        \stepcounter{imcnt}
      }
    }
\end{tikzpicture}%

    \caption{Jacobian columns for a $\beta$-VAE with 25 latent dimensions} 
  \end{subfigure}
\medskip

\noindent 
  \setcounter{imcnt}{0}
  \begin{subfigure}[]{12cm}
    \centering
  \def \gammaval{10}%
  \begin{tikzpicture}
    \foreach \y in {0,...,4} {
      \foreach \x in {0,...,4} {
        \node[inner sep=0] at (\x, -\y)
         {\includegraphics[width=0.9cm]{latentJacobianImages/naturalImageNorm/beta0_010_ica0_0\gammaval_lat25_im0_latind\theimcnt_x\ximcropstart_y\yimcropstart}};
        \stepcounter{imcnt}
      }
    }
  \end{tikzpicture}
    \caption{Jacobian columns for a JL1-VAE with 25 latent dimensions} 
  \end{subfigure}
  \caption{Results for $\beta$-VAE and JL1-VAE using 25 latent dimensions
  (instead of the 10 shown in the main paper). Both are trained with
  $\beta=0.01$, and JL1-VAE trained with $\gamma=0.01$.
  }
  \label{fig:additionalNaturalImageLatents25}
\end{figure}

\section{MPI3D-Multi Experiment Hyperparameters}
We train a $\beta$-VAE with $\beta = 0.01$ 
on a 2$\times$2 tiling of every-other-pixel downsampling of randomly sampled images pulled from
MPI3D-real,
resulting in 
64$\times$64-pixel training images.
We only sample MPI3D-real images of a top-down view of the robot holding a
large, blue cube, with salmon background lighting.
In this way, our dataset does not vary along unordered/sparse latent factors like color/shape.
This leaves two independent dimensions of variance (horizontal and vertical axis
joints) for each of the 4 tiled robot images.
If we were to apply our local disentanglement metrics to discrete factors of variation that do not come with a natural  distance metric such as ``shape'' or ``color,'' we would require equality in order for values to be considered ``close.''
That is, for any such factor of variation, a ``local'' sampled dataset will be constant on that factor.

We embed the dataset into a latent space of 10 dimensions.
We train on 300,000 independently sampled batches of 64 images from the cache, giving a total of
19,200,000 image presentations to the neural network.
The $\beta$ was chosen to give good reconsruction accuracy.
Additionally, we train our JL1-VAE with the same $\beta$ and model architecture on the same training dataset
with our added $L_1$ regularization weighted by a hyperparameter $\gamma = 0.01$.
The hyperparameter $\gamma$ was chosen as the largest tested for which the learning algorithm converged to give good reconstruction accuracy.
We use linear annealing for both the $\beta$ and $\gamma$ parameters, annealing
each from 0 to their final values over the first 100,000 batches.
We use the Adam optimizer with a learning rate of 0.0001.

\ifthenelse{\equal{\detokenize{nohyperref}}{\jobname}}{
\clearpage
\section*{Checklist}

\begin{enumerate}

\item For all authors...
\begin{enumerate}
  \item Do the main claims made in the abstract and introduction accurately reflect the paper's contributions and scope?
    \answerYes{}
  \item Did you describe the limitations of your work?
    \answerYes{See Section~\ref{sec:Discussion}}
  \item Did you discuss any potential negative societal impacts of your work?
    \answerNA{The authors were not able to identify potential negative societal impacts of this work.}
  \item Have you read the ethics review guidelines and ensured that your paper conforms to them?
    \answerYes{}
\end{enumerate}

\item If you are including theoretical results...
\begin{enumerate}
  \item Did you state the full set of assumptions of all theoretical results?
    \answerNA{}
	\item Did you include complete proofs of all theoretical results?
    \answerNA{}
\end{enumerate}

\item If you ran experiments...
\begin{enumerate}
  \item Did you include the code, data, and instructions needed to reproduce the main experimental results (either in the supplemental material or as a URL)?
    \answerYes{}
  \item Did you specify all the training details (e.g., data splits, hyperparameters, how they were chosen)?
    \answerYes{}
	\item Did you report error bars (e.g., with respect to the random seed after running experiments multiple times)?
    \answerYes{Figure~\ref{fig:localMetricVaryingRho} and Figure~\ref{fig:localMetricComparedToStandardMethods} show results for training each model ten times with different random seeds}
	\item Did you include the total amount of compute and the type of resources used (e.g., type of GPUs, internal cluster, or cloud provider)?
    \answerYes{}
\end{enumerate}

\item If you are using existing assets (e.g., code, data, models) or curating/releasing new assets...
\begin{enumerate}
  \item If your work uses existing assets, did you cite the creators?
    \answerYes{}
  \item Did you mention the license of the assets?
    \answerYes{}
  \item Did you include any new assets either in the supplemental material or as a URL?
    \answerYes{}
  \item Did you discuss whether and how consent was obtained from people whose data you're using/curating?
    \answerNA{}
  \item Did you discuss whether the data you are using/curating contains personally identifiable information or offensive content?
    \answerNA{}
\end{enumerate}

\item If you used crowdsourcing or conducted research with human subjects...
\begin{enumerate}
  \item Did you include the full text of instructions given to participants and screenshots, if applicable?
    \answerNA{}
  \item Did you describe any potential participant risks, with links to Institutional Review Board (IRB) approvals, if applicable?
    \answerNA{}
  \item Did you include the estimated hourly wage paid to participants and the total amount spent on participant compensation?
    \answerNA{}
\end{enumerate}

\end{enumerate}
}{} 

\end{document}